%% file: main.tex
\documentclass[runningheads]{llncs}

\usepackage{eccv}

\usepackage{eccvabbrv}

\usepackage{multirow}
\usepackage{graphicx}
\usepackage{booktabs}
\usepackage{amssymb}
\usepackage{colortbl}
\usepackage{color}
\usepackage{indentfirst}
\definecolor{green1}{RGB}{26,153,25}
\definecolor{red1}{RGB}{220,20,1}
\usepackage{enumitem}
\usepackage[accsupp]{axessibility}  

\usepackage[pagebackref,breaklinks,colorlinks,citecolor=eccvblue]{hyperref}

\begin{document}

\title{Region-aware Distribution Contrast: \\A Novel Approach to Multi-Task \\Partially Supervised Learning} 

\titlerunning{Region-aware Distribution Contrast}

\author{Meixuan Li \and
Tianyu Li \and
Guoqing Wang\inst{*} \and \\ 
Peng Wang \and Yang Yang \and Heng Tao Shen}

\authorrunning{M.~Li et al.}

\institute{University of Electronic Science and Technology of China (UESTC)  \\
\email{limeixuan0801@126.com, \{cosmos.yu, p.wang6, shenhengtao\}@hotmail.com, \\ 
\{gqwang0420, yang.yang\}@uestc.edu.cn}\\
}

\maketitle

\begin{abstract}
  In this study, we address the intricate challenge of multi-task dense prediction, encompassing tasks such as semantic segmentation, depth estimation, and surface normal estimation, particularly when dealing with partially annotated data (MTPSL). The complexity arises from the absence of complete task labels for each training image. Given the inter-related nature of these pixel-wise dense tasks, our focus is on mining and capturing cross-task relationships. Existing solutions typically rely on learning global image representations for global cross-task image matching, imposing constraints that, unfortunately, sacrifice the finer structures within the images. Attempting local matching as a remedy faces hurdles due to the lack of precise region supervision, making local alignment a challenging endeavor. The introduction of Segment Anything Model (SAM) sheds light on addressing local alignment challenges by providing free and high-quality solutions for region detection. Leveraging SAM-detected regions, the subsequent challenge lies in aligning the representations within these regions. Diverging from conventional methods that directly learn a monolithic image representation, our proposal involves modeling region-wise representations using Gaussian Distributions. Aligning these distributions between corresponding regions from different tasks imparts higher flexibility and capacity to capture intra-region structures, accommodating a broader range of tasks. This innovative approach significantly enhances our ability to effectively capture cross-task relationships, resulting in improved overall performance in partially supervised multi-task dense prediction scenarios. Extensive experiments conducted on two widely used benchmarks underscore the superior effectiveness of our proposed method, showcasing state-of-the-art performance even when compared to fully supervised methods.
\end{abstract}

\vspace{-3em}
\section{Introduction}
\label{sec:intro}
\vspace{-0.5em}
\indent With the rapid advancement of deep learning, dense prediction tasks, including semantic segmentation, depth estimation, and surface normal estimation, have witnessed remarkable progress \cite{eigen2014depth,he2017mask,long2015fully, poggi2020uncertainty,zhang2020uc}. Considering the inherent interdependence among these dense prediction tasks \cite{zamir2020robust,zamir2018taskonomy}, there has been a growing interest in employing unified multi-task learning networks to jointly tackle different dense prediction tasks \cite{liu2019mtan,lu2017fullymtl,zhang2019patternmtl,misra2016cross,xu2018padnet,vandenhende2020mtinet, chen2018gradnorm, liu2022auto,kendall2018uncertainty}. In contrast to inefficient single-task learning networks, multi-task learning networks effectively acquire shared features, enabling them to capture more generalizable information across various tasks while avoiding redundant training \cite{vandenhende2020mtinet,zhang2019patternmtl,xu2018padnet}. 

\input{Fig_text/Fig1}
Recent advancements in multi-task dense prediction methods have primarily focused on two crucial aspects. The first aspect is dedicated to designing intricate network architectures to achieve effective multi-task learning \cite{misra2016cross,xu2018padnet,liu2019mtan,vandenhende2020mtinet,zhang2019patternmtl}, while the second aspect is devoted to devising optimal balancing strategies for loss functions in multi-task learning \cite{liu2019mtan,chen2018gradnorm,liu2022auto,kendall2018uncertainty,guo2018dynamic}, aiming to mitigate the negative transfer across tasks. These directions mainly focus on fully supervised multi-task learning, where complete task labels are available for all training samples. However, acquiring pixel-level labels for dense prediction tasks is resource-intensive, particularly with multiple visual tasks. It is common to encounter scenarios where labels for certain tasks are missing or unreliable in some training samples \cite{li2022mtpsl}, leading to a new research direction known as multi-task partially supervised learning (MTPSL).

Since not all task labels are available for each training image in MTPSL, it is natural to formulate a multi-task learning framework which can leverage the implicit relationship between tasks with and without explicit labels. Dense prediction tasks, such as semantic segmentation, depth estimation, and normal prediction, exhibit a complementary co-existence relationship \cite{zamir2020robust,zamir2018taskonomy,standley2020tasks}, where each task can serve as auxiliary information for others, offering valuable insights reciprocally. While existing approaches map predictions into a joint vector-wise semantic space for task alignment~\cite{li2022mtpsl}, we argue that this overlooks the underlying structural information inherent in the original image. A global vector might be insufficient to characterize the information in the entire scene, limiting its effectiveness in constraining cross-task relationships. 
Local matching, as shown in~\cref{fig:fig1}, with its intrinsic local properties, serves as a potential remedy. Nonetheless, the absence of precise region supervision presents a fundamental obstacle, thereby intensifying the challenges associated with its implementation.

To tackle the outlined challenges, we introduce region-based cross-task alignment by leveraging the robust region segmentation capabilities inherent in Segment Anything Model (SAM) models. Utilizing well-segmented image regions facilitates the abstraction of region-level information, enabling the addition of cross-task consistency constraints. However, a subsequent crucial challenge arises in effectively representing these regional features to ensure the consistency and accuracy of cross-task alignment. 
In contrast to conventional methods that directly learn a unified image representation, we employ Gaussian distributions to model region representations, enhancing flexibility and improving the ability to capture intra-region structures. By adapting Gaussian distributions between corresponding regions from different tasks, our approach can be more broadly applicable to diverse task types, such as pixel-level classification tasks (e.g., semantic segmentation) and pixel-level regression tasks (e.g., depth estimation).
Subsequently, we introduce distributional contrastive learning to explicitly perform region-wise cross-task alignment. This approach serves as a robust constraint, offering a meaningful way to mine relationships between tasks and enhance dense prediction results, particularly for those tasks lacking pixel-wise labels.

In summary, our work masks the following contributions:
\begin{itemize}
    \item We tackle the challenge of multi-task partially supervised (MTPSL) dense prediction from a fresh perspective. Our approach involves extracting cross-task local alignment through the utilization of SAM's easily obtainable local regions. This innovative strategy has demonstrated notable in alleviating label shortages.
    \item We propose a novel region distribution contrast method for local alignment, which offers increased flexibility, robustness, and wide applicability across multiple tasks.
    \item We validate the effectiveness of our proposed method through extensive experiments conducted on two widely used benchmark datasets. Notably, our approach achieves state-of-the-art performance in partially supervised learning, while also showing great potential in fully supervised learning. 
\end{itemize}
\vspace{-1em}
\section{Related Work}
\label{sec:rel_work}
\vspace{-1em}
\textbf{Multi-task Supervised Learning.} Multi-task learning utilizes a single model to simultaneously handle multiple visual tasks, offering advantages such as faster inference speed and more efficient utilization of input data. Some approaches \cite{misra2016cross,gao2019nddr,zhang2022automtl} focus on designing the encoder of multi-task network, while other approaches pay more attention to designing complex decoders to generate task-specific predictions by utilizing shared features \cite{xu2018padnet, vandenhende2020mtinet,zhang2019patternmtl,bruggemann2021exploring}. Uncertainty \cite{kendall2018uncertainty} obtain task loss weights by considering the heteroscedastic uncertainty of each task, Grad-Norm \cite{chen2018gradnorm} directly adjusts gradients to balance task losses, and DWA \cite{guo2018dynamic} learns time-varying average task weights by considering the loss variation rate of each task. It is important to note that these methods operate under a significant and strong assumption, namely that they are developed and studied based on the availability of labels for all tasks across all images.

\textbf{Multi-task Semi-supervised Learning and Partially-supervised Learning.} Learning multi-task models on fully annotated data requires a large-scale labeled dataset, and the cost of collecting sufficient labeled data can be high. Therefore, two common scenarios are worthy of note in multi-task learning. One is semi-supervised multi-task learning, where the dataset consists of limited annotation information for all tasks and a large amount of unlabeled data. A bunch of methods have been proposed for semi-supervised multi-task learning \cite{liu2007semi, chen2020multi,huang2020partly,latif2020multi,imran2019semi}. 
In recent works \cite{chen2020multi,huang2020partly,latif2020multi,imran2019semi}, regularization terms are applied to unlabeled samples from each task to encourage consistent predictions when the input is perturbed. Another scenario involves partial-supervision multi-task learning, where the dataset lacks sufficient labels for each task, and not all tasks have labels for every image. Li~\etal~\cite{li2022mtpsl} maps tasks into high-dimensional vectors for cross-task alignment. However, vector representations are insufficient and do not capture local-level alignment. In this paper, we propose leveraging distributions as a substitute for vector representations of task features and employing contrastive learning to achieve region-level alignment across tasks.

\textbf{Contrastive Learning for Vision.} Recently, contrastive learning has made significant advancements in unsupervised learning \cite{oord2018representation,wu2018unsupervised, chen2020simple,chen2020improved, he2020momentum}. DenseCL \cite{wang2021dense} and RegionContrast \cite{hu2021region} have been utilized for semantic segmentation tasks using contrastive learning at the pixel and region levels respectively, achieving excellent results. For depth estimation, WCL \cite{fan2023depthcontrastive} transformed depth from continuous values to discrete values and constructed contrastive losses based on windows to form positive and negative sample pairs. CMC \cite{tian2020contrastive} was the first to propose that different viewpoints of the same image should be mapped to nearby positions in a high-dimensional space, while different images should be mapped to distant positions, thus ensuring multiview consistency. Inspired by CMC, we believe that multiple viewpoints within the same local region should also exhibit consistency, while differing from other regions. In this paper, we leverage this idea to achieve region-level contrast for MTPSL.

\textbf{Gaussian Distribution.} 
Recently, Gaussian distribution has been widely employed in computer vision to characterize the distribution of features \cite{zhang2023noisy, liang2022gmmseg, jin2020global, chen2021hsva, chang2020data, wu2023sparsely}. GMMSeg \cite{liang2022gmmseg} utilized the Expectation-Maximization algorithm to construct Gaussian Mixture Models for each class in semantic segmentation, capturing class-conditional densities. AGMM \cite{wu2023sparsely} constructed an adaptive Gaussian Mixture Model for sparse annotation in semantic segmentation by incorporating labeled pixels and their similar unlabeled counterparts. DUL \cite{chang2020data} employed mean and variance to characterize the distribution of faces for face recognition. Gaussian distributions have the capability to probabilistically describe the distribution characteristics of different visual tasks. However, to the best of our knowledge, Gaussian distributions have not been applied in MTPSL.
\vspace{-0.2cm}

\section{Method}
\subsection{Preliminaries}
\textbf{Problem Setup.} Considering the problem of multi-task learning involving $K$ tasks, where $K \geq 2$. A training dataset $S=\left\{{x_{i}}\right\}^{N}_{i=1}$ is given with $N$ partially labelled samples, indicating that for each training sample $x_i$, only a subset of the $K$ tasks are provided true labels.
Let $P_i$ represents the number of labeled tasks for $x_i$ and $Q_i$ represents the number of unlabeled tasks, such that $P_i+Q_i=K$. When $P_i$ equals $K$ for all $x_i \in S$, it indicates fully supervised multi-task learning. Conversely, when $Q_i$ equals $K$ for all $x_i \in S$, it implies that no task labels are available, leading to unsupervised multi-task learning. In this paper, we focus on addressing partially supervised multi-task learning, where each training image $x_i$ can obtain labels for at least one task ($P_i \geq 1$).
\vspace{-0.2cm}

\subsection{Overall of Our Method}
\input{Fig_text/Fig_Framework}
As shown in~\cref{fig:framework}, our method focuses on utilizing Gaussian distributions to represent features in local regions for contrastive learning, achieving better distribution consistency across tasks at the region-level for MTPSL. Assuming that task $s$ has label while task $t$ does not, our objective is to achieve alignment between the true label $\mathbf{\mathit{y}}_s$ of task $s$ and the prediction $\hat{y}_t$ of task $t$ at the region level using contrastive learning.

In the training stage, to accurately and efficiently obtain region-level features for prediction, we simultaneously feed the input image to both the multi-task learning (MTL) network and the pre-trained SAM. Predictions are obtained from both MTL network and SAM. For tasks with labeled data, we apply fully-supervised loss $L_{Sup}$ as explicit training supervision. For tasks without label, an auxiliary network $a_\theta$ maps the unlabelled task prediction $\hat{y}_t$ and the annotated task label $y_s$ to the same high-dimensional feature map space, and use $L_{RC}$ to achieve cross-task consistency at the region level. Specifically, We first extract region-wise features in $\hat{y}_t$ and $y_s$ based on the segmented regions identified by SAM, then model them as Gaussian distributions. After acquiring the Gaussian distributions for features in each region of each task, we utilize contrastive learning to minimize the distance between the Gaussian distributions of features in the same region across different tasks and maximize the distance from the Gaussian distributions of features from different regions. 
Finally, the optimization of our method can be defined as:
\begin{equation}
    Loss = L_{Sup} + L_{RC}.
\end{equation}

In the inference stage, as illustrated in~\cref{fig:framework}, the input image is fed only into the MTL network to generate predictions for multiple tasks, notably, with SAM no longer being utilized. 
\vspace{-0.2cm}
\subsection{Region-aware Contrastive Learning}
\input{Fig_text/Fig3}
In this section, we explain the implementation details to achieve region-level cross-task consistency.

\textbf{Region-level Features Extraction.} To realize the region-level cross-task consistency, it is essential to informatively choose regions for feature extraction. Considering the robust capabilities of SAM for image segmentation \cite{Kirillov2023sam}, we employ it to generate the regions for each training sample, leveraging its ability to provide valuable prior knowledge. As shown in~\cref{fig:framework}, to obtain fine-grained semantics and fine-grained edges of regions, we fed the image into MTL network and pre-trained SAM simultaneously. SAM performs fine-grained segmentation of each object in the image, and its output masks contain the ID of each area and its corresponding position in the image. Inspired by \cite{jin2023let}, for convenient region-wise feature extraction in task-specific feature maps, we compute the predicted masks of SAM as a grayscale image, where each pixel value corresponds to the area ID returned by SAM, indicating its category. Leveraging the grayscale masks of SAM, we split the image into different regions and extract region-wise features within feature maps of different tasks. 

\textbf{Gaussian Distribution Modeling.} After obtaining the region-wise features for each task, we consider modeling the region-wise features as Gaussian distributions. The utilization of Gaussian distributions represents the features of a region in a probabilistic manner, which provides a more comprehensive depiction of the region's variations and compensates for the shortcomings of pixel-level hard alignment. Specifically, we model each region $r_i$ from the task-specific feature map $f_t$ as a Gaussian distribution:
\begin{eqnarray}
    p_{i,t} = p(r_i|f_t) = \mathit{N} (r_i;\boldsymbol\mu_{i,t}, \boldsymbol{\Sigma}_{i,t}),
\end{eqnarray}
where $\boldsymbol\mu_{i,t}$ and $\boldsymbol{\Sigma}_{i,t}$ represent mean and covariance matrix respectively, calculated from the region $r_i$ in the feature map $f_t$, as shown in~\cref{fig:fig3}. 

\textbf{Region Distribution Contrast.} Once we have obtained the Gaussian distributions for each region, our objective is to utilize contrastive learning to minimize the distance between Gaussian distributions of the same region across different tasks and maximize the distance between Gaussian distributions of different regions. Hence, the region-wise Gaussian distribution contrastive loss for the region distribution $p_{i,t}$ can be defined as follows:
\begin{equation}
    L_{i,t}^{N C E}=-\log \frac{sim \left(p_{i,t} , k_{+} \right)}{sim \left(p_{i,t} , k_{+} \right)+\sum_{k_{-}} sim \left(p_{i,t} , k_{-}\right)},
\end{equation}
\begin{equation}
    sim(p_{i,t} , k) = exp(-W_{distance}(p_{i,t}, k) / \tau),
\end{equation}
where $k^+$ and $k^-$ respectively represent the positive and negative samples of $p_{i,t}$, $k$ denotes $k^+$ or $k^-$, $sim$ denotes the exponential equation of the Wasserstein distance, and $\tau$ is the temperature parameter. Wasserstein distance \cite{ruschendorf1985wasserstein} is a metric used to quantify the dissimilarity between two probability distributions and well-suited for accurately measuring the distance between discrete and continuous distributions, as it accounts for the underlying structure and geometry of the data. It measures the minimum amount of work required to transform one distribution into another and can be calculated as follows:
\begin{equation}
    \begin{aligned}
    W_{distance}(p_{i,t}, k)&=\left\|\boldsymbol\mu_{i,t}-\boldsymbol{\mu}_{k}\right\|^{2}+\operatorname{Tr}\left(\boldsymbol{\Sigma}_{i,t}\right) \\ &+\operatorname{Tr}\left(\boldsymbol{\Sigma}_{k}\right)-2 \operatorname{Tr}\left(\left(\boldsymbol{\Sigma}_{i,t} \boldsymbol{\Sigma}_{k}\right)^{1 / 2}\right)
    \end{aligned},
\end{equation}
where $\operatorname{Tr}$ represents the trace of the matrix, $\boldsymbol{\mu}_{k}$ and $\boldsymbol{\Sigma}_{k}$ represent the mean and covariance matrix of sample $k$.
Based on $L^{N C E}$, the region-wise contrastive loss $L_{RC}$ can be defined as follows:
\begin{equation}
    L_{RC} = \frac{1}{KM} \sum_{t}^{K}\sum_{i}^{M} L_{i,t}^{N C E},
\end{equation}
where $M$ is the number of regions in the feature map $t$.

By minimizing $L_{RC}$, the network learns to contrast the distributions from the same region with those from different regions, thereby enabling better cross-task consistency at the region level.

\textbf{Alternative Local Extraction Strategy.}
Here we explore an alternative local extraction strategy to investigate local cross-task consistency. To extract local information from the feature map, the simplest approach is to divide the map into patches, where patches at the same position serve as positive samples, and their distances are minimized. Conversely, patches from other positions act as negative samples, and their distances are maximized. We refer to this local extraction strategy as patch-aware contrast, which will be compared to the region-aware contrast method in~\cref{experiments}.

\textbf{Alternative Local Contrast Strategy.}
Alternatively, the method of modeling regions as Gaussian distributions and performing contrastive learning can be replaced with other contrastive strategies. One strategy is pixel-to-pixel level region contrast, where each pixel within the region is pulled closer to the corresponding pixel in the positive sample region and pushed far apart from each pixel in the negative sample region. We refer to this as pixel contrast. Another strategy is to map the region to a vector and bring it closer to the vector of the positive sample region while pushing it away from the vector of the negative sample region. This strategy is referred to as vector contrast. We compare these alternative methods in~\cref{experiments}.
\vspace{-0.2cm}
\section{Experiments}
\label{experiments}
\input{Tab_text/nyuv2}
\textbf{Datasets.} We evaluated our method on two standard dense prediction multi-task datasets: NYU-V2 \cite{silberman2012indoor} and Cityscapes \cite{cordts2016cityscapes}. NYU-V2 is an indoor dataset of natural scenes, including three dense prediction tasks: semantic segmentation, depth estimation, and surface normal estimation. The semantic segmentation task in NYU-V2 contains 13 categories, depth estimation is provided by Microsoft Kinect, and surface normal estimation is provided by \cite{eigen2015predicting}. As \cite{liu2019mtan, li2022mtpsl}, all images were resized to 288x384 for training and evaluation purposes. Cityscapes is a well-known dataset for autonomous driving, consisting of outdoor street scenes with two tasks: semantic segmentation and depth estimation. Following \cite{liu2019mtan, li2022mtpsl}, we used 7-class semantic segmentation annotations for evaluation and resized all images to 128x256 to improve training speed. 

\textbf{Experimental Setting.} Following \cite{li2022mtpsl}, we evaluated our method in two partially supervised multi-task settings: the one label setting and the random label setting. In the one label setting, each image is assigned a random task with a label. In the random label setting, each image is assigned random labels for multiple tasks, with at least one task having a label and at most $K-1$ tasks having labels, where $K$ represents the total number of tasks in the multi-task setup.

\textbf{Implementation Details and Evaluation Metrics.}
All our experiments were conducted on the NVIDIA A100 GPU. For NYU-V2, we trained the models for 20 epochs, while for Cityscapes, we trained them for 100 epochs. Following \cite{liu2019mtan,li2022mtpsl}, we employed cross-entropy loss for semantic segmentation, L1 loss for depth estimation, and cosine similarity loss for surface normal estimation. We utilized the exact same evaluation metrics as mentioned in \cite{liu2019mtan, li2022mtpsl}. For the semantic segmentation task, we employed the mean intersection over union (mIoU) metric. The absolute error (aErr) was used to evaluate the depth estimation task, and the mean error (mErr) was employed for surface normal estimation.
\subsection{Quantitative Evaluation}
\input{Tab_text/cityscapes}
\input{Tab_text/nyu_abla}
\input{Fig_text/Fig_onelabel}
\input{Fig_text/fig4}
\textbf{Quantitative Results on NYU-V2 Dataset.} The evaluation results are shown in~\cref{exp:nyu}. Due to the absence of pre-trained parameters in Li~\etal~\cite{li2022mtpsl}, the training process exhibits slower convergence. To expedite the convergence speed, we replaced the backbone of Li~\etal~\cite{li2022mtpsl} with HRNet18, which possesses a smaller training parameter size but utilizes pre-trained weights. The comparisons between Li~\etal~\cite{li2022mtpsl} with SegNet as the backbone and Li~\etal~\cite{li2022mtpsl} with HRNet18 as the backbone are presented in the first row and the fourth row, respectively, for each label setting. Since the source code of DejaVu \cite{borse2023dejavu} is not publicly available, we are unable to replace its backbone with HRNet18 for a fair comparison. To resolve this issue, we provide the results of our method based on SegNet, which utilizes the same backbone. In all the methods compared using the same backbone, our approach demonstrates significant advantages. In the one label setting, the patch-aware Gaussian contrast method demonstrates significant improvements across all three tasks compared to Li~\etal~\cite{li2022mtpsl} across all three tasks, as it achieves a certain degree of local-level cross-task alignment. Furthermore, the region-aware Gaussian contrast method further enhances performance, achieving improvements of 2.48\%, 2.25\% and 5.51\%  over the SOTA~\cite{borse2023dejavu} across the three tasks, owing to the achievement of cross-task consistency at the object level. In the random label setting, the superiority of the region-aware Gaussian contrast method can be observed, as it outperforms other methods significantly across all three tasks, yielding improvements of 4.70\%, 4.18\% and 2.85\%, respectively, over the SOTA~\cite{borse2023dejavu}.

Furthermore, we provide a comparison with the state-of-the-art fully supervised method, TaskExpert~\cite{ye2023taskexpert}.~\cref{exp:nyu} clearly shows that our method significantly improves the performance of TaskExpert in partially supervised scenarios. Particularly in the random setting, when combined with our method, TaskExpert achieves even better performance than the fully supervised TaskExpert in the surface normal task. This once again demonstrates the effectiveness and wide applicability of our approach.

\textbf{Quantitative Results on Cityscapes Dataset.} The evaluation results are describe in~\cref{exp:cityscapes}. Similar to NYU-V2, we first compared Li~\etal~\cite{li2022mtpsl} with HRNet18 as the backbone. Subsequently, we discussed the results of the region-aware contrast method under different contrast strategies. It is observed that Gaussian contrast demonstrates improvements in semantic segmentation and depth estimation compared to vector-based methods, indicating that representing regions using Gaussian distributions provides a more comprehensive and accurate representation than using vectors. In terms of computational speed, the Gaussian contrast method is faster and more efficient, while still performing on par with pixel-based contrast methods.
\vspace{-1em}
\subsection{Qualitative Evaluation}
The qualitative evaluation of onelabel setting, random label setting on the NYU-V2 dataset, and onelabel setting on the Cityscapes dataset are shown in~\cref{fig:fig_onelabel},~\cref{fig:fig4} and~\cref{fig:fig5}, respectively. We present a comparative analysis of the results from two sets of images for each setting of each dataset. As indicated by~\cref{fig:fig_onelabel} and~\cref{fig:fig4}, it can be observed that our method exhibits superior accuracy and smoothness in semantic segmentation, which is particularly evident in~\cref{fig:fig4}. Regarding depth estimation, our method, which is based on region-aware Gaussian distribution contrast, achieves more accurate predictions when objects undergo changes compared to Li~\etal~\cite{li2022mtpsl}. The latter fails to capture such variations effectively. As for surface normal estimation, our method outperforms Li~\etal~\cite{li2022mtpsl} overall and demonstrates finer contour handling. These observations highlight the advantages of our approach over Li~\etal~\cite{li2022mtpsl} in terms of classification accuracy, smoothness in semantic segmentation, accurate depth estimation in the presence of object changes, and finer handling of surface normals.

We further exhibit the multi-task prediction results on the Cityscapes dataset in~\cref{fig:fig5}. Benefited from our proposed region-aware distribution contrast learning method, both the semantic segmentation and depth estimation tasks exhibit improved capability in capturing object edges and fine-grained details compared to Li~\etal~\cite{li2022mtpsl}. Additionally, there are significant overall performance improvement at a global level. 
\vspace{-1em}

\input{Tab_text/SAM_contribution}
\subsection{Ablation Study}
\label{abla}
\input{Fig_text/fig5}
We conduct ablation study on NYU-V2 dataset to prove the effectiveness of our method from various aspects.

\textbf{Local Extraction Strategy.} ~\cref{exp:nyu_abla} provides a comparison between different local extraction strategies. All methods are compared based on HRNet18 as the backbone. It can be observed that in both the onelabel setting and random label setting, regardless of the contrast strategy employed, the region-aware contrast method significantly outperforms the patch-aware contrast method. This is because the region-aware method takes into account the objects present in the image and performs contrast based on objects, rather than simply comparing patches that may contain multiple objects.

\textbf{Local Contrast Strategy.} ~\cref{exp:nyu_abla} also provides a comparison between different local contrast strategies. All methods are similarly compared based on HRNet18. It can be observed that regardless of the label setting and the local extraction strategy employed, the Gaussian distribution contrast consistently outperforms the Vector contrast and approaches or even surpasses the pixel contrast. This is because in multi-task learning scenarios that involve both discrete and continuous dense predictions, utilizing Gaussian distributions to represent regions for contrast allows for better cross-task alignment in the overall distribution, whereas the hard alignment strategy of pixel contrast may lead to inaccuracies.

\textbf{Contribution of SAM.} As shown in~\cref{exp:cityscapes},~\cref{exp:nyu_abla}, and~\cref{exp:full} in our article, although the vector contrast approach also incorporates SAM, it fails to yield satisfactory results. In contrast, the adoption of distribution contrast exhibits a significant improvement over the vector contrast approach. To further demonstrate this point, we provide a comparison of results using semantic segmentation labels instead of SAM in fully supervised setting, as shown in~\cref{exp:sam_contribution}. This further emphasizes that solely introducing fine-grained segmentation results is insufficient, highlighting the importance of the distribution contrast method.

\input{Tab_text/dist}
\textbf{Distance Measurement for Distributions.}~\cref{exp:dist} presents a comparison of different distance measurements. While KL divergence is commonly used to calculate the distance between Gaussian distributions, it fails to accurately reflect the distance between discrete and continuous distributions, which is crucial in the context of multi-task alignment. On the other hand, Wasserstein distance has the advantage of measuring the distance between two non-overlapping distributions. The experimental results in~\cref{exp:dist} demonstrate that Wasserstein distance is better suited for evaluating the distance between different task distributions. More detailed information can be found in the \textcolor{red!60!black}{supplementary materials}.

\input{Tab_text/full}
\textbf{Fully Supervised Setting.} ~\cref{exp:full} presents the comparison of our method in the fully supervised setting. All methods are based on HRNet18. Compared to other methods, the region-aware distribution contrast method once again demonstrates superiority, outperforming significantly in all three tasks. This is because Gaussian distributions can characterize the distribution characteristics of different tasks and facilitate accurate and efficient alignment with both vector and pixel-based approaches.
\vspace{-1em}
\section{Conclusion}
\label{sec:conclusion}
In this paper, we propose a novel region-aware distribution contrast learning method for MTPSL. To facilitate cross-task alignment at the region level, predictions from different tasks are initially mapped into a joint feature map space. Simultaneously, SAM is employed to generate regions for each training sample, leveraging its ability to provide valuable prior knowledge. Subsequently, region-wise features are modeled by fitting a Gaussian distribution. The alignment among regions is achieved by minimizing the disparity among distributions from different tasks within the same region and maximizing the distance between distributions from different regions. The effectiveness of our method is demonstrated on two standard multi-task datasets through extensive experiments, outperforming the state-of-the-art by a significant margin. Although our method is designed for MTPSL, it can be applied to general multi-task learning problems to leverage the intrinsic relationships across different tasks, as demonstrated in the fully supervised experiments in~\cref{abla}. Moreover, we believe that the proposed region-aware distribution contrast learning method provides a promising way for solving more general multi-modal problems, and we will investigate this in our future work.


%
%
\clearpage
\bibliographystyle{splncs04}
\bibliography{main}

\end{document}

%% file: Fig_text/Fig1.tex
\begin{figure}[!t]
\centering
\includegraphics[width=1.0\linewidth]
{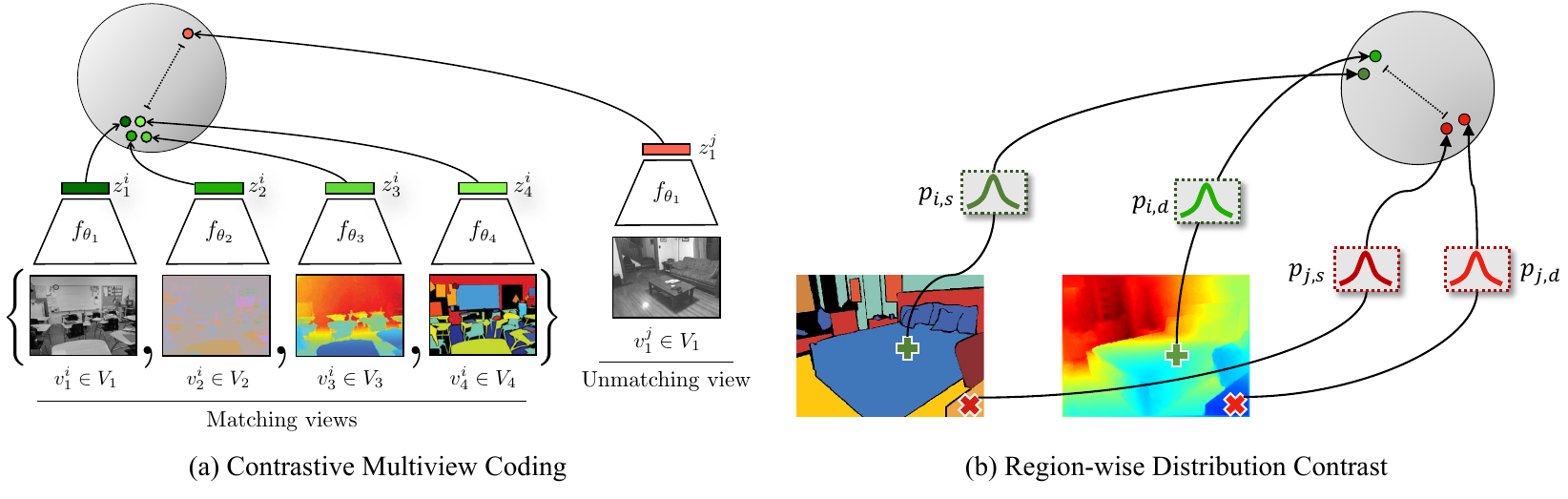}
\vspace{-2em}
\caption{\small{Multiview consistency and region consistency. (a) Illustration of contrastive multiview consistency \cite{tian2020contrastive}. (b) Illustration of region consistency, where $p_{i,s}$ and $p_{j,s}$ represent the region distribution of semantic segmentation, $p_{i,d}$ and $p_{j,d}$ denotes the region distribution of depth estimation.}}
\vspace{-2.5em}
\label{fig:fig1}

\end{figure}

%% file: Fig_text/Fig_Framework.tex
\begin{figure*}[!t]
\centering
\setlength{\belowcaptionskip}{-0.45cm}
\includegraphics[width=1.075\linewidth]
{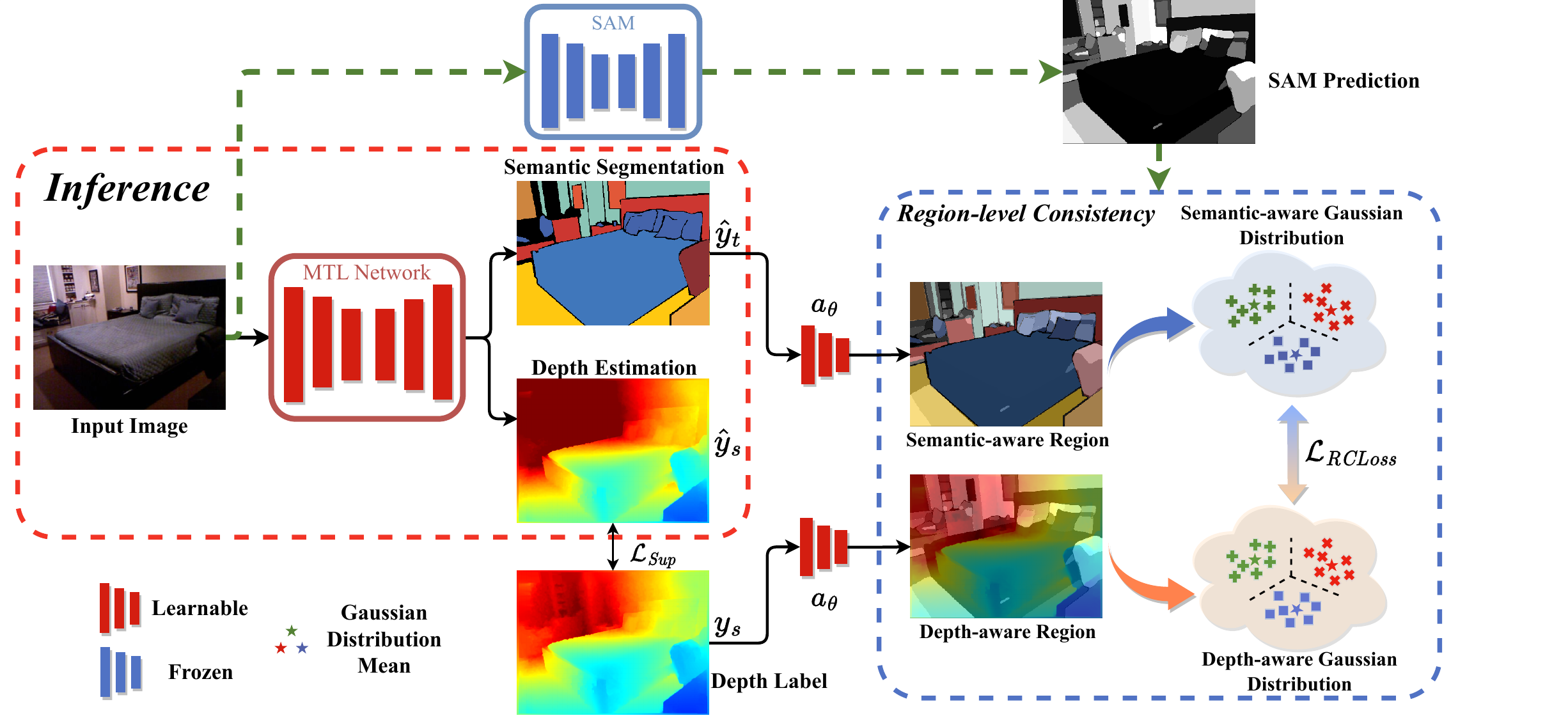}

\caption{\small{Illustration of our region-aware distribution contrast learning method for MTPSL. During training, supervised constraints $L_{Sup}$ are applied to the annotated task $s$. For task $s$ and unlabelled task $t$, the $a_{\theta}$ map the true label $y_s$ and the prediction $\hat{y}_t$ in the high-dimensional space respectively and then model the region of task-specific features extracted using SAM as a Gaussian distribution. Contrastive learning is then employed to minimize the distance between Gaussian distributions of the same region across different tasks, while maximizing the distance to Gaussian distributions of other regions.}}
\label{fig:framework}

\end{figure*}

%% file: Fig_text/Fig3.tex
\begin{figure*}[!t]
\centering
\setlength{\belowcaptionskip}{-0.45cm}
\includegraphics[width=1.0\linewidth]{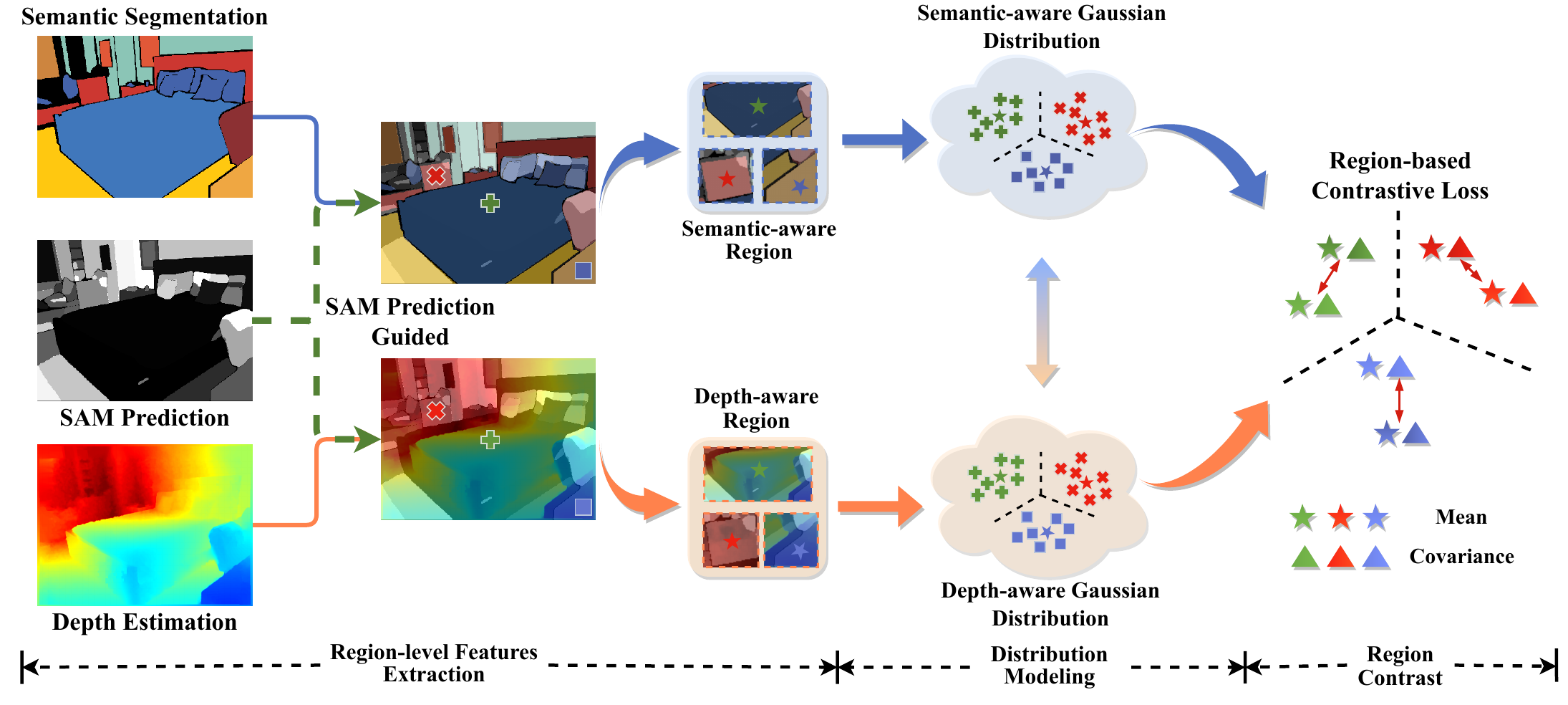}

\caption{\small{Illustration of region-level cross-task consistency. Initially, SAM predictions are employed to extract regions from the features of the particular task. Following that, these regions are modeled as Gaussian distributions. Finally, contrastive learning is utilized to minimize the distance between regions of the same region across different tasks and maximize the distance to regions of other regions.}}
\label{fig:fig3}

\end{figure*}

%% file: Tab_text/nyuv2.tex
\begin{table}[!t]
\caption{\textbf{Quantitative comparison on the NYU-V2 dataset.} $\uparrow$ ($\downarrow$) denotes that, larger (smaller) values lead to better quality. ${\textcolor{red}\ast}$ represents the available partially supervised SOTA. ${\textcolor{green}\ast}$ represents the available fully supervised SOTA. \small \textcolor{red1}{+} denotes the percentage \textcolor{red1}{improvement} over the SOTA performance, corresponding to $\uparrow$ ($\downarrow$). The bold denotes the best.}
\vspace{-1em}
\label{exp:nyu}
\centering
\resizebox{\linewidth}{!}{%
\begin{tabular}{@{}ccc||cc||ccc@{}}
\toprule
             &                &     & \multicolumn{2}{c||}{\textbf{Extraction strategy}} & \multicolumn{3}{c}{\textbf{Tasks}}                 \\ \cmidrule(l){4-8}
\multirow{-2}{*}{\textbf{Label setting}} &
  \multirow{-2}{*}{\textbf{Method}} &
  \multirow{-2}{*}{\textbf{Backbone}} &
  \textbf{Patch} &
  \textbf{Region} &
  \textbf{Semantic(mIoU)}$\uparrow $ &
  \textbf{Depth(aErr)} $\downarrow $&
  \textbf{Normals(mErr)}$\downarrow $ \\
   \midrule
full     & TaskExpert$^{\textcolor{green}\ast}$~\cite{ye2023taskexpert}             &       ViT-Large &             &                        & 57.58         & 0.3730          & 5.90        \\
\midrule \midrule
onelabel     & Li~\etal~\cite{li2022mtpsl}             &       SegNet &             &                        & 30.36          & 0.6088          & 32.08      \\
onelabel     & DejaVu$^{\textcolor{red}\ast}$ \cite{borse2023dejavu}            &    SegNet &                    &                        & 31.02          & 0.5959          & 32.15          \\
\rowcolor[HTML]{EFEFEF}
onelabel     & Gaussian Contrast(Ours)            &     SegNet &                   &    \checkmark                    & 31.79\,\footnotesize \textcolor{red1}{(+2.48)}         & 0.5835\,\footnotesize \textcolor{red1}{(+2.25)}          & 30.38\,\footnotesize \textcolor{red1}{(+5.51)}           \\ 
\midrule
onelabel     & Li~\etal$^{\textcolor{red}\ast}$~\cite{li2022mtpsl}            &       HRNet18 &             &                        & 39.42          & 0.5071          & 14.79           \\
onelabel     & DejaVu \cite{borse2023dejavu}            &    HRNet18 &                    &                        & —          & —          & —           \\
onelabel     & Gaussian Contrast &   HRNet18 & \checkmark            &                        & 41.56\,\footnotesize \textcolor{red1}{(+5.43)}          & 0.4884\,\footnotesize \textcolor{red1}{(+3.69)}          & 7.71\,\footnotesize \textcolor{red1}{(+47.87)}          \\
\rowcolor[HTML]{C0C0C0} 
onelabel     & Gaussian Contrast(Ours) &   HRNet18  &                      &      \checkmark             & \textbf{42.28\,\footnotesize \textcolor{red1}{(+7.26)}} & \textbf{0.4641\,\footnotesize \textcolor{red1}{(+8.48)}} & \textbf{4.86\,\footnotesize \textcolor{red1}{(+67.14)}} \\ 
\midrule
onelabel     & TaskExpert$^{\textcolor{green}\ast}$~\cite{ye2023taskexpert}             &       ViT-Large &             &                        & 49.53         & 0.4305          & 11.12           \\
onelabel     & TaskExpert~\cite{ye2023taskexpert}+Ours            &       ViT-Large &             &       \checkmark        & 55.81\,\footnotesize \textcolor{red1}{(+12.68)}         & 0.4092\,\footnotesize \textcolor{red1}{(+4.95)}          & 8.46\,\footnotesize \textcolor{red1}{(+23.92)}           \\
\midrule \midrule
random label & Li~\etal~\cite{li2022mtpsl} &   SegNet            &                &                        & 34.26          & 0.5787          & 31.06           \\
random label & DejaVu$^{\textcolor{red}\ast}$ \cite{borse2023dejavu}  &   SegNet          &                        &                        & 35.72          & 0.5665          & 29.82           \\
\rowcolor[HTML]{EFEFEF}
random label & Gaussian Contrast(Ours)   &   SegNet         &                        &          \checkmark              & 37.40\,\footnotesize \textcolor{red1}{(+4.70)}          & 0.5428\,\footnotesize \textcolor{red1}{(+4.18)}          & 28.97\,\footnotesize \textcolor{red1}{(+2.85)}           \\ \midrule
random label & Li~\etal$^{\textcolor{red}\ast}$~\cite{li2022mtpsl}  &   HRNet18            &                &                        & 41.35          & 0.4845          & 14.34           \\
random label  & DejaVu \cite{borse2023dejavu}            &    HRNet18 &                    &                        & —          & —          & —           \\
random label & Gaussian Contrast &   HRNet18 &      \checkmark          &                        & 45.79\,\footnotesize \textcolor{red1}{(+10.74)}          & 0.4619\,\footnotesize \textcolor{red1}{(+4.66)}          & 7.37\,\footnotesize \textcolor{red1}{(+48.61)}          \\
\rowcolor[HTML]{C0C0C0} 
random label & Gaussian Contrast(Ours) &   HRNet18 &                      &      \checkmark        & \textbf{46.21\,\footnotesize \textcolor{red1}{(+11.75)}} & \textbf{0.4482\,\footnotesize \textcolor{red1}{(+7.49)}} & \textbf{4.49\,\footnotesize \textcolor{red1}{(+68.69)}} \\ 
\midrule 
random label     & TaskExpert$^{\textcolor{green}\ast}$~\cite{ye2023taskexpert}             &       ViT-Large &             &                        & 56.32         & 0.4282          & 6.47           \\
random label     & TaskExpert~\cite{ye2023taskexpert}+Ours            &       ViT-Large &             &       \checkmark        & 57.14\,\footnotesize \textcolor{red1}{(+1.46)}         & 0.3977\,\footnotesize \textcolor{red1}{(+7.12)}          & 4.78\,\footnotesize \textcolor{red1}{(+26.12)}           \\
\bottomrule
\end{tabular}%
}
\end{table}

%% file: Tab_text/cityscapes.tex
\begin{table*}[!t]
\caption{\textbf{Quantitative comparison on the Cityscapes dataset.} $\uparrow$ ($\downarrow$) denotes that, larger (smaller) values lead to better quality. ${\textcolor{red}\ast}$ represents the available SOTA. \small \textcolor{red1}{+} denotes the percentage \textcolor{red1}{improvement} over the SOTA performance, corresponding to $\uparrow$ ($\downarrow$). The bold denotes the best.}
\vspace{-1em}
\label{exp:cityscapes}
\centering
\resizebox{\linewidth}{!}{%
\begin{tabular}{@{}ccc||ccc||cc@{}}
\toprule
         &                    &  &\multicolumn{3}{c||}{\textbf{Contrast strategy}} &        
         \multicolumn{2}{c}{\textbf{Tasks}} \\ \cmidrule(l){4-8}
\multirow{-2}{*}{\textbf{Label setting}} &
  \multirow{-2}{*}{\textbf{Method}} &
  \multirow{-2}{*}{\textbf{Backbone}} &
  \textbf{Vector} &
  \textbf{Pixel} &
  \textbf{Gaussian} &
  \textbf{Semantic(mIoU)}$\uparrow $ &
  \textbf{Depth(aErr)} $\downarrow $ \\  \midrule
onelabel & Li~\etal~\cite{li2022mtpsl} & SegNet               &        &          &           & 74.90           & 0.0161           \\
onelabel & Li~\etal$^{\textcolor{red}\ast}$~\cite{li2022mtpsl}  & HRNet18               &        &          &           & 81.73           & 0.0157           \\ \midrule
onelabel & Region-aware Contrast & HRNet18 &   \checkmark     &               &                & 82.76\,\footnotesize \textcolor{red1}{(+1.26)}           & 0.0145\,\footnotesize \textcolor{red1}{(+7.64)}           \\
onelabel & Region-aware Contrast & HRNet18 &       & \checkmark   &                & 83.20\,\footnotesize \textcolor{red1}{(+1.80)}         & 0.0141\,\footnotesize \textcolor{red1}{(+10.19)}           \\
\rowcolor[HTML]{EFEFEF} 
onelabel & Region-aware Contrast (Ours) & HRNet18 &        &               &   \checkmark   & \textbf{83.92\,\footnotesize \textcolor{red1}{(+2.68)}}           & \textbf{0.0121\,\footnotesize \textcolor{red1}{(+22.73)}}           \\   \bottomrule
\end{tabular}%
}
\end{table*}

%% file: Tab_text/nyu_abla.tex
\begin{table*}[t]
\caption{Ablation study on the NYU-V2 dataset to explore the performance of different local extraction strategies and different local contrast strategies.}
\vspace{-1em}
\label{exp:nyu_abla}
\centering
\resizebox{\linewidth}{!}{%
\begin{tabular}{@{}c||cc||ccc||ccc@{}}
\toprule
\multirow{2}{*}{\textbf{Label setting}} & \multicolumn{2}{c||}{\textbf{Extraction strategy}} & \multicolumn{3}{c||}{\textbf{Contrast strategy}} & \multicolumn{3}{c}{\textbf{Tasks}} \\ \cmidrule(l){2-9} 
 & \textbf{Patch} & \textbf{Region} & \textbf{Vector} & \textbf{Pixel} & \textbf{Gaussian} & \textbf{Semantic(mIoU)}$\uparrow $ & \textbf{Depth(aErr)}$\downarrow $ & \textbf{Normals(mErr)}$\downarrow $ \\ \midrule
onelabel & \checkmark &  & \checkmark &  &  & 40.24 & 0.4998 & 7.9518 \\
onelabel & \checkmark &  &  & \checkmark &  & 41.50 & 0.4934 & 8.3648 \\
\rowcolor[HTML]{EFEFEF} 
onelabel & \checkmark &  &  &  & \checkmark & 41.56 & 0.4884 & 7.7141 \\ \midrule
onelabel &  & \checkmark & \checkmark &  &  & 41.84 & 0.4727 & 5.8098 \\
onelabel &  & \checkmark &  & \checkmark &  & 42.05 & 0.4663 & 4.9150 \\
\rowcolor[HTML]{C0C0C0} 
onelabel &  & \checkmark &  &  & \checkmark & \textbf{42.28} & \textbf{0.4641} & \textbf{4.8613} \\ \midrule \midrule
random label & \checkmark &  & \checkmark &  &  & 43.98 & 0.4795 & 7.4464 \\
random label & \checkmark &  &  & \checkmark &  & 44.62 & 0.4777 & 7.8873 \\
\rowcolor[HTML]{EFEFEF} 
random label & \checkmark &  &  &  & \checkmark & 45.79 & 0.4619 & 7.3747 \\ \midrule
random label &  & \checkmark & \checkmark &  &  & 45.83 & 0.4630 & 4.8393 \\
random label &  & \checkmark &  & \checkmark &  & \textbf{46.41} & 0.4518 & 4.8442 \\
\rowcolor[HTML]{C0C0C0} 
random label &  & \checkmark &  &  & \checkmark & 46.21 & \textbf{0.4482} & \textbf{4.4907} \\ \bottomrule
\end{tabular}%
}
\end{table*}

%% file: Fig_text/Fig_onelabel.tex
\begin{figure*}[!ht]
\centering
\setlength{\belowcaptionskip}{-0.40cm}
\setlength{\abovecaptionskip}{-0.05cm}
\includegraphics[width=1.0\linewidth]{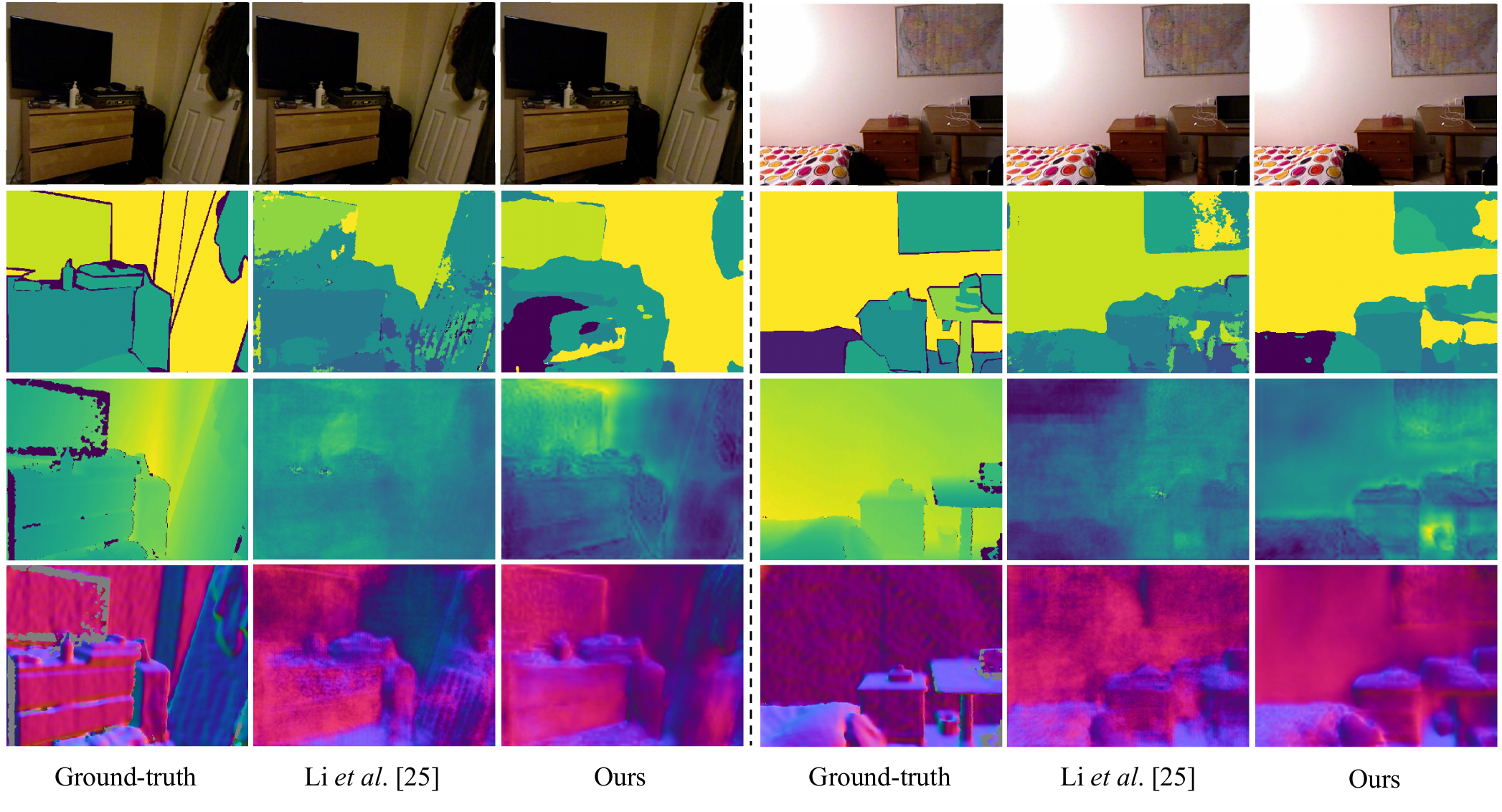}
\caption{\small{\textbf{Qualitative results of onelabel setting on NYU-V2.} The first row shows the input image, the second row represents the ground-truth or predictions of semantic segmentation, the third row plots the ground-truth or predictions of depth estimation, and the final row presents the ground-truth or predictions of surface normal estimation. }}
\label{fig:fig_onelabel}

\end{figure*}

%% file: Fig_text/Fig4.tex
\begin{figure*}[!ht]
\centering
\setlength{\belowcaptionskip}{-0.65cm}
\setlength{\abovecaptionskip}{-0.05cm}
\includegraphics[width=1.0\linewidth]{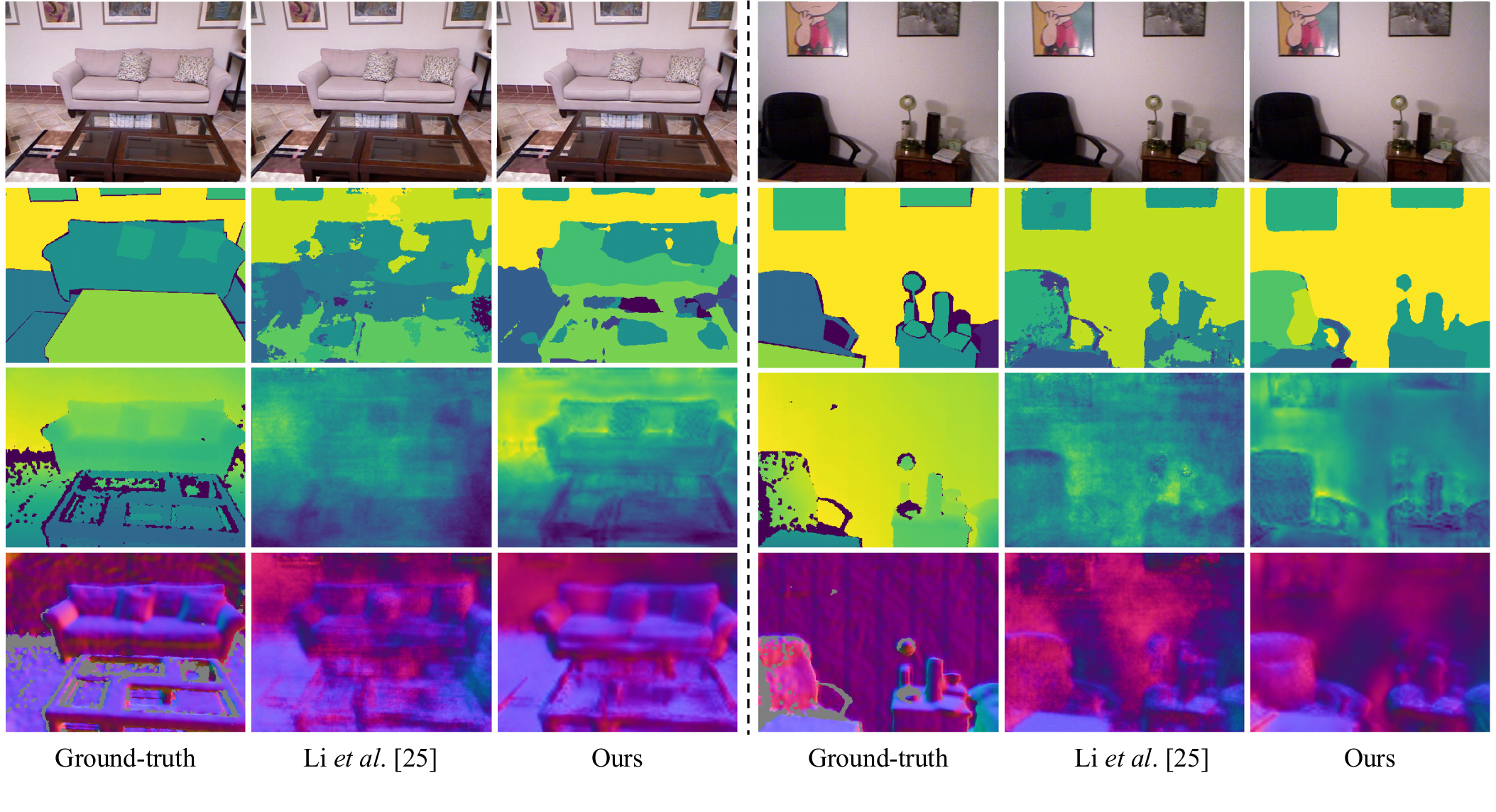}
\caption{\small{\textbf{Qualitative results of random label setting on NYU-V2.} The first row shows the input image, the second row represents the ground-truth or predictions of semantic segmentation, the third row plots the ground-truth or predictions of depth estimation, and the final row presents the ground-truth or predictions of surface normal estimation. }}
\label{fig:fig4}

\end{figure*}

%% file: Tab_text/SAM_contribution.tex
\begin{table}[h]
\caption{Ablation study on the NYU-V2 dataset to investigate the contribution of SAM in the \textbf{fully supervised setting}.}
\vspace{-1em}
\label{exp:sam_contribution}
\centering
\resizebox{\linewidth}{!}{%
\begin{tabular}{@{}c|cc||ccc@{}}
\toprule
 {\textbf{Method}} & \textbf{Vector} &   \textbf{Gaussian} & \textbf{Semantic(mIoU)}$\uparrow $ & \textbf{Depth(aErr)}$\downarrow $ & \textbf{Normals(mErr)}$\downarrow $\\ \midrule
 Region-aware Contrast & \checkmark &  & 49.00 & 0.4206 & 8.1878 \\
\rowcolor[HTML]{EFEFEF} 
 Region-aware Contrast (Ours) &  &  \checkmark & \textbf{51.14} & \textbf{0.3982} & \textbf{4.8108} \\ \bottomrule
\end{tabular}%
}
\end{table}
\vspace{-3em}

%% file: Fig_text/Fig5.tex
\begin{figure*}[!t]
\setlength{\belowcaptionskip}{-0.45cm}
\centering
\includegraphics[width=1.0\linewidth]{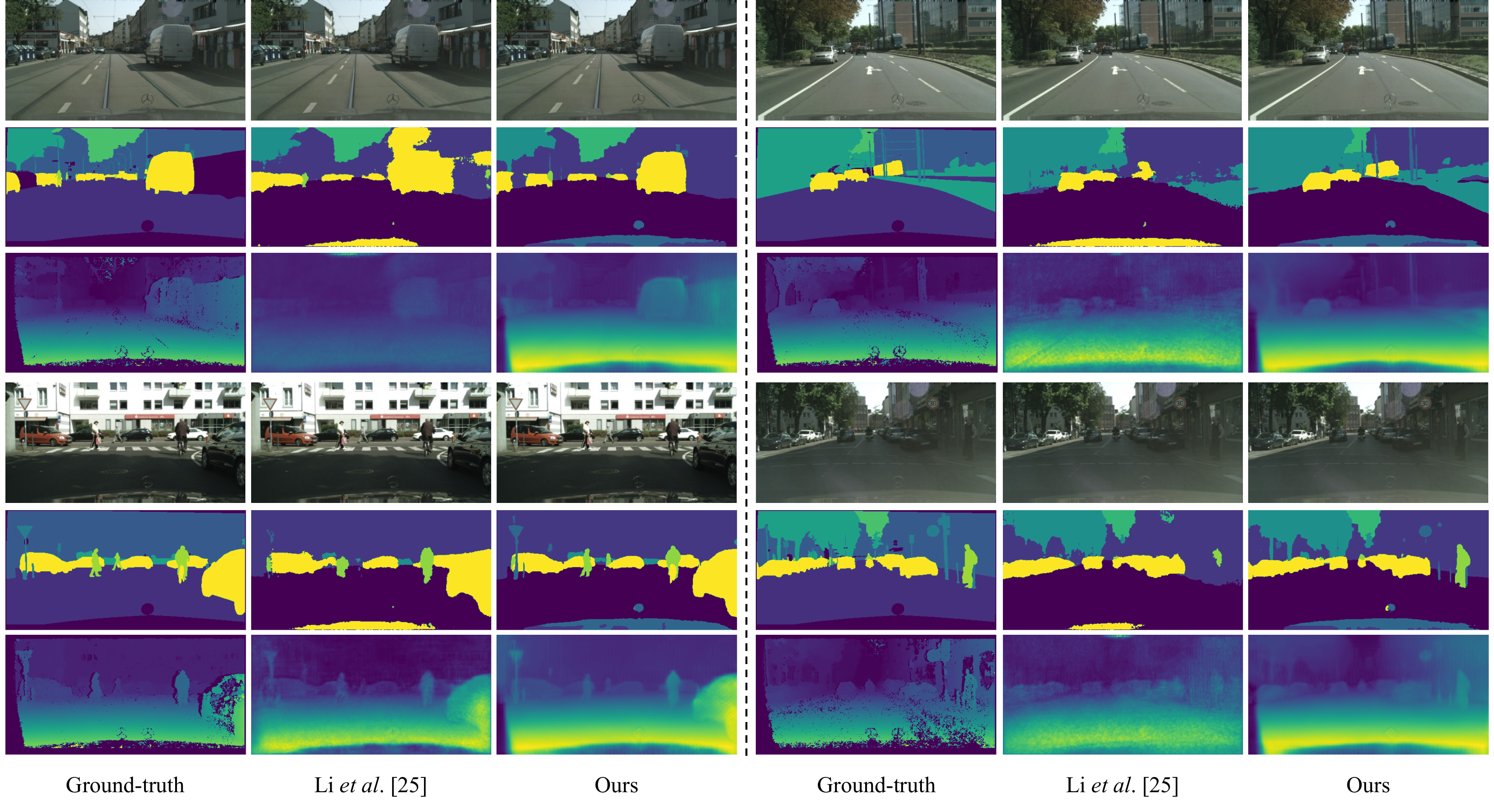}
\caption{\small{\textbf{Qualitative results of onelabel setting on Cityscapes.} The first row shows the input image, the second row represents the ground-truth or predictions of semantic segmentation, and the final row plots the ground-truth or predictions of depth estimation. }}
\label{fig:fig5}

\end{figure*}

%% file: Tab_text/dist.tex
\begin{table}[t]
\caption{Ablation study on the NYU-V2 dataset to investigate the performance of various distance measurements.}
\vspace{-1em}
\label{exp:dist}
\centering
\resizebox{\columnwidth}{!}{%
\begin{tabular}{@{}cc||ccc@{}}
\toprule
    & & \multicolumn{3}{c}{\textbf{Tasks}} \\ \cmidrule(l){3-5} 
   \multirow{-2}{*}{\textbf{Label setting}} &
  \multirow{-2}{*}{\textbf{Method}} & \textbf{Semantic(mIoU)}$\uparrow $ & \textbf{Depth(aErr)}$\downarrow $ & \textbf{Normals(mErr)}$\downarrow $\\ \midrule
 onelabel & KL divergence &  41.15 & 0.4739 & 4.94 \\
\rowcolor[HTML]{EFEFEF}
 onelabel & Wasserstein distance (Ours) &  \textbf{42.28} & \textbf{0.4641} & \textbf{4.86} \\ 
\midrule
random label & KL divergence &  45.53 & 0.4539 & 5.04 \\
\rowcolor[HTML]{EFEFEF}
random label & Wasserstein distance (Ours) &  \textbf{46.21} & \textbf{0.4482} & \textbf{4.49} \\ 
\bottomrule
\end{tabular}%
}
\end{table}


%% file: Tab_text/full.tex
\begin{table}[t]
\setlength{\abovecaptionskip}{-0.25cm}
\caption{Ablation study on the NYU-V2 dataset to investigate the performance of various methods in the \textbf{fully supervised setting}.}
\vspace{-1em}
\label{exp:full}
\centering
\resizebox{\columnwidth}{!}{%
\begin{tabular}{@{}c||ccc||ccc@{}}
\toprule
  & \multicolumn{3}{c||}{\textbf{Contrast strategy}} & \multicolumn{3}{c}{\textbf{Tasks}} \\ \cmidrule(l){2-7} 
  \multirow{-2}{*}{\textbf{Method}} & \textbf{Vector} & \textbf{Pixel} & \textbf{Gaussian} & \textbf{Semantic(mIoU)}$\uparrow $ & \textbf{Depth(aErr)}$\downarrow $ & \textbf{Normals(mErr)}$\downarrow $\\ \midrule
 MTL &  &  &  & 46.38 & 0.4660 & 6.7298 \\
 Li~\etal~\cite{li2022mtpsl} &  &  &  & 47.20 & 0.4477 & 5.1192 \\
 Region-aware Contrast & \checkmark &  &  & 47.64 & 0.4231 & 5.1072 \\
 Region-aware Contrast &  & \checkmark &  & 50.06 & 0.4168 & 4.5318 \\
\rowcolor[HTML]{EFEFEF} 
 Region-aware Contrast (Ours) &  &  & \checkmark & \textbf{50.83} & \textbf{0.4051} & \textbf{3.2878} \\ \bottomrule
\end{tabular}%
}
\end{table}


%% file: main.bbl
\begin{thebibliography}{10}
\providecommand{\url}[1]{\texttt{#1}}
\providecommand{\urlprefix}{URL }
\providecommand{\doi}[1]{https://doi.org/#1}

\bibitem{borse2023dejavu}
Borse, S., Das, D., Park, H., Cai, H., Garrepalli, R., Porikli, F.: Dejavu: Conditional regenerative learning to enhance dense prediction. In: Proceedings of the IEEE/CVF Conference on Computer Vision and Pattern Recognition. pp. 19466--19477 (2023)

\bibitem{bruggemann2021exploring}
Br{\"u}ggemann, D., Kanakis, M., Obukhov, A., Georgoulis, S., Van~Gool, L.: Exploring relational context for multi-task dense prediction. In: Proceedings of the IEEE/CVF International Conference on Computer Vision. pp. 15869--15878 (2021)

\bibitem{chen2020simple}
Chen, T., Kornblith, S., Norouzi, M., Hinton, G.: A simple framework for contrastive learning of visual representations. In: International Conference on Machine Learning. pp. 1597--1607. PMLR (2020)

\bibitem{chen2020improved}
Chen, X., Fan, H., Girshick, R., He, K.: Improved baselines with momentum contrastive learning. arXiv preprint arXiv:2003.04297  (2020)

\bibitem{chen2018gradnorm}
Chen, Z., Badrinarayanan, V., Lee, C.Y., Rabinovich, A.: {GradNorm}: Gradient normalization for adaptive loss balancing in deep multi-task networks. In: International Conference on Machine Learning. pp. 794--803. PMLR (2018)

\bibitem{chen2020multi}
Chen, Z., Zhu, L., Wan, L., Wang, S., Feng, W., Heng, P.A.: A multi-task mean teacher for semi-supervised shadow detection. In: Proceedings of the IEEE/CVF Conference on Computer Vision and Pattern Recognition. pp. 5611--5620 (2020)

\bibitem{cordts2016cityscapes}
Cordts, M., Omran, M., Ramos, S., Rehfeld, T., Enzweiler, M., Benenson, R., Franke, U., Roth, S., Schiele, B.: The cityscapes dataset for semantic urban scene understanding. In: Proceedings of the IEEE Conference on Computer Vision and Pattern Recognition. pp. 3213--3223 (2016)

\bibitem{deng2011hierarchical}
Deng, J., Berg, A.C., Fei-Fei, L.: Hierarchical semantic indexing for large scale image retrieval. In: Proceedings of the IEEE Conference on Computer Vision and Pattern Recognition. pp. 785--792. IEEE (2011)

\bibitem{douze2011combining}
Douze, M., Ramisa, A., Schmid, C.: Combining attributes and fisher vectors for efficient image retrieval. In: Proceedings of the IEEE Conference on Computer Vision and Pattern Recognition. pp. 745--752. IEEE (2011)

\bibitem{eigen2015predicting}
Eigen, D., Fergus, R.: Predicting depth, surface normals and semantic labels with a common multi-scale convolutional architecture. In: Proceedings of the IEEE International Conference on Computer Vision. pp. 2650--2658 (2015)

\bibitem{eigen2014depth}
Eigen, D., Puhrsch, C., Fergus, R.: Depth map prediction from a single image using a multi-scale deep network. Advances in Neural Information Processing Systems  \textbf{27} (2014)

\bibitem{fan2023depthcontrastive}
Fan, R., Poggi, M., Mattoccia, S.: Contrastive learning for depth prediction. In: Proceedings of the IEEE/CVF Conference on Computer Vision and Pattern Recognition. pp. 3225--3236 (2023)

\bibitem{gao2019nddr}
Gao, Y., Ma, J., Zhao, M., Liu, W., Yuille, A.L.: Nddr-cnn: Layerwise feature fusing in multi-task cnns by neural discriminative dimensionality reduction. In: Proceedings of the IEEE/CVF Conference on Computer Vision and Pattern Recognition. pp. 3205--3214 (2019)

\bibitem{guo2018dynamic}
Guo, M., Haque, A., Huang, D.A., Yeung, S., Fei-Fei, L.: Dynamic task prioritization for multitask learning. In: Proceedings of the European Conference on Computer Vision. pp. 270--287 (2018)

\bibitem{he2020momentum}
He, K., Fan, H., Wu, Y., Xie, S., Girshick, R.: Momentum contrast for unsupervised visual representation learning. In: Proceedings of the IEEE/CVF Conference on Computer Vision and Pattern Recognition. pp. 9729--9738 (2020)

\bibitem{he2017mask}
He, K., Gkioxari, G., Doll{\'a}r, P., Girshick, R.: Mask r-cnn. In: Proceedings of the IEEE International Conference on Computer Vision. pp. 2961--2969 (2017)

\bibitem{hu2021region}
Hu, H., Cui, J., Wang, L.: Region-aware contrastive learning for semantic segmentation. In: Proceedings of the IEEE/CVF International Conference on Computer Vision. pp. 16291--16301 (2021)

\bibitem{huang2020partly}
Huang, C., Tang, H., Fan, W., Xiao, Y., Hao, D., Qian, Z., Terzopoulos, D., et~al.: Partly supervised multi-task learning. In: 2020 19th IEEE International Conference on Machine Learning and Applications. pp. 769--774. IEEE (2020)

\bibitem{imran2019semi}
Imran, A.A.Z., Terzopoulos, D.: Semi-supervised multi-task learning with chest x-ray images. In: Machine Learning in Medical Imaging: 10th International Workshop, MLMI 2019, Held in Conjunction with MICCAI 2019, Shenzhen, China, October 13, 2019, Proceedings 10. pp. 151--159. Springer (2019)

\bibitem{jin2020global}
Jin, X., Lan, C., Zeng, W., Chen, Z.: Global distance-distributions separation for unsupervised person re-identification. In: Proceedings of the European Conference on Computer Vision. pp. 735--751. Springer (2020)

\bibitem{jin2023let}
Jin, Z., Chen, S., Chen, Y., Xu, Z., Feng, H.: Let segment anything help image dehaze. arXiv preprint arXiv:2306.15870  (2023)

\bibitem{kendall2018uncertainty}
Kendall, A., Gal, Y., Cipolla, R.: Multi-task learning using uncertainty to weigh losses for scene geometry and semantics. In: Proceedings of the IEEE Conference on Computer Vision and Pattern Recognition. pp. 7482--7491 (2018)

\bibitem{Kirillov2023sam}
Kirillov, A., Mintun, E., Ravi, N., Mao, H., Rolland, C., Gustafson, L., Xiao, T., Whitehead, S., Berg, A.C., Lo, W.Y., Dollar, P., Girshick, R.: Segment anything. In: Proceedings of the IEEE/CVF International Conference on Computer Vision. pp. 4015--4026 (October 2023)

\bibitem{latif2020multi}
Latif, S., Rana, R., Khalifa, S., Jurdak, R., Epps, J., Schuller, B.W.: Multi-task semi-supervised adversarial autoencoding for speech emotion recognition. IEEE Transactions on Affective Computing  \textbf{13}(2),  992--1004 (2020)

\bibitem{li2022mtpsl}
Li, W.H., Liu, X., Bilen, H.: Learning multiple dense prediction tasks from partially annotated data. In: Proceedings of the IEEE/CVF Conference on Computer Vision and Pattern Recognition. pp. 18879--18889 (2022)

\bibitem{liang2022gmmseg}
Liang, C., Wang, W., Miao, J., Yang, Y.: Gmmseg: Gaussian mixture based generative semantic segmentation models. Advances in Neural Information Processing Systems  \textbf{35},  31360--31375 (2022)

\bibitem{liu2007semi}
Liu, Q., Liao, X., Carin, L.: Semi-supervised multitask learning. Advances in Neural Information Processing Systems  \textbf{20} (2007)

\bibitem{liu2022auto}
Liu, S., James, S., Davison, A., Johns, E.: Auto-lambda: Disentangling dynamic task relationships. Transactions on Machine Learning Research  (2022)

\bibitem{liu2019mtan}
Liu, S., Johns, E., Davison, A.J.: End-to-end multi-task learning with attention. In: Proceedings of the IEEE/CVF Conference on Computer Vision and Pattern Recognition. pp. 1871--1880 (2019)

\bibitem{long2015fully}
Long, J., Shelhamer, E., Darrell, T.: Fully convolutional networks for semantic segmentation. In: Proceedings of the IEEE Conference on Computer Vision and Pattern Recognition. pp. 3431--3440 (2015)

\bibitem{lu2017fullymtl}
Lu, Y., Kumar, A., Zhai, S., Cheng, Y., Javidi, T., Feris, R.: Fully-adaptive feature sharing in multi-task networks with applications in person attribute classification. In: Proceedings of the IEEE Conference on Computer Vision and Pattern Recognition. pp. 5334--5343 (2017)

\bibitem{misra2016cross}
Misra, I., Shrivastava, A., Gupta, A., Hebert, M.: {Cross-Stitch} networks for multi-task learning. In: Proceedings of the IEEE Conference on Computer Vision and Pattern Recognition. pp. 3994--4003 (2016)

\bibitem{oord2018representation}
Oord, A.v.d., Li, Y., Vinyals, O.: Representation learning with contrastive predictive coding. arXiv preprint arXiv:1807.03748  (2018)

\bibitem{poggi2020uncertainty}
Poggi, M., Aleotti, F., Tosi, F., Mattoccia, S.: On the uncertainty of self-supervised monocular depth estimation. In: Proceedings of the IEEE/CVF Conference on Computer Vision and Pattern Recognition. pp. 3227--3237 (2020)

\bibitem{ruschendorf1985wasserstein}
R{\"u}schendorf, L.: The wasserstein distance and approximation theorems. Probability Theory and Related Fields  \textbf{70}(1),  117--129 (1985)

\bibitem{silberman2012indoor}
Silberman, N., Hoiem, D., Kohli, P., Fergus, R.: Indoor segmentation and support inference from rgbd images. In: Proceedings of the European Conference on Computer Vision. pp. 746--760. Springer (2012)

\bibitem{standley2020tasks}
Standley, T., Zamir, A., Chen, D., Guibas, L., Malik, J., Savarese, S.: Which tasks should be learned together in multi-task learning? In: International Conference on Machine Learning. pp. 9120--9132. PMLR (2020)

\bibitem{tian2020contrastive}
Tian, Y., Krishnan, D., Isola, P.: Contrastive multiview coding. In: Proceedings of the European Conference on Computer Vision. pp. 776--794. Springer (2020)

\bibitem{vandenhende2020mtinet}
Vandenhende, S., Georgoulis, S., Van~Gool, L.: {MTI-Net}: Multi-scale task interaction networks for multi-task learning. In: Proceedings of the European Conference on Computer Vision. pp. 527--543. Springer (2020)

\bibitem{wang2021exploring}
Wang, W., Zhou, T., Yu, F., Dai, J., Konukoglu, E., Van~Gool, L.: Exploring cross-image pixel contrast for semantic segmentation. In: Proceedings of the IEEE/CVF International Conference on Computer Vision. pp. 7303--7313 (2021)

\bibitem{wang2021dense}
Wang, X., Zhang, R., Shen, C., Kong, T., Li, L.: Dense contrastive learning for self-supervised visual pre-training. In: Proceedings of the IEEE/CVF Conference on Computer Vision and Pattern Recognition. pp. 3024--3033 (2021)

\bibitem{wu2023sparsely}
Wu, L., Zhong, Z., Fang, L., He, X., Liu, Q., Ma, J., Chen, H.: Sparsely annotated semantic segmentation with adaptive gaussian mixtures. In: Proceedings of the IEEE/CVF Conference on Computer Vision and Pattern Recognition. pp. 15454--15464 (2023)

\bibitem{wu2018unsupervised}
Wu, Z., Xiong, Y., Yu, S.X., Lin, D.: Unsupervised feature learning via non-parametric instance discrimination. In: Proceedings of the IEEE Conference on Computer Vision and Pattern Recognition. pp. 3733--3742 (2018)

\bibitem{xu2018padnet}
Xu, D., Ouyang, W., Wang, X., Sebe, N.: {PAD-Net}: Multi-tasks guided prediction-and-distillation network for simultaneous depth estimation and scene parsing. In: Proceedings of the IEEE Conference on Computer Vision and Pattern Recognition. pp. 675--684 (2018)

\bibitem{ye2023taskexpert}
Ye, H., Xu, D.: Taskexpert: Dynamically assembling multi-task representations with memorial mixture-of-experts. In: Proceedings of the IEEE/CVF International Conference on Computer Vision. pp. 21828--21837 (2023)

\bibitem{zamir2020robust}
Zamir, A.R., Sax, A., Cheerla, N., Suri, R., Cao, Z., Malik, J., Guibas, L.J.: Robust learning through cross-task consistency. In: Proceedings of the IEEE/CVF Conference on Computer Vision and Pattern Recognition. pp. 11197--11206 (2020)

\bibitem{zamir2018taskonomy}
Zamir, A.R., Sax, A., Shen, W., Guibas, L.J., Malik, J., Savarese, S.: Taskonomy: Disentangling task transfer learning. In: Proceedings of the IEEE Conference on Computer Vision and Pattern Recognition. pp. 3712--3722 (2018)

\bibitem{zhang2020uc}
Zhang, J., Fan, D.P., Dai, Y., Anwar, S., Saleh, F.S., Zhang, T., Barnes, N.: Uc-net: Uncertainty inspired rgb-d saliency detection via conditional variational autoencoders. In: Proceedings of the IEEE/CVF Conference on Computer Vision and Pattern Recognition. pp. 8582--8591 (2020)

\bibitem{zhang2022automtl}
Zhang, L., Liu, X., Guan, H.: Automtl: A programming framework for automating efficient multi-task learning. Advances in Neural Information Processing Systems  \textbf{35},  34216--34228 (2022)

\bibitem{zhang2023noisy}
Zhang, M., Zhao, X., Yao, J., Yuan, C., Huang, W.: When noisy labels meet long tail dilemmas: A representation calibration method. In: Proceedings of the IEEE/CVF International Conference on Computer Vision. pp. 15890--15900 (2023)

\bibitem{zhang2019patternmtl}
Zhang, Z., Cui, Z., Xu, C., Yan, Y., Sebe, N., Yang, J.: Pattern-affinitive propagation across depth, surface normal and semantic segmentation. In: Proceedings of the IEEE/CVF Conference on Computer Vision and Pattern Recognition. pp. 4106--4115 (2019)

\end{thebibliography}
